# A KAN-based Interpretable Framework for Process-Informed Prediction of Global Warming Potential

Jaewook Lee, Xinyang Sun‡, Ethan Errington‡, Miao Guo*

Department of Engineering, King's College London, Strand, London, UK

## Abstract

Accurate prediction of Global Warming Potential (GWP) is essential for assessing the environmental impact of chemical processes and materials. Traditional GWP prediction models rely predominantly on molecular structure, overlooking critical process-related information. In this study, we present an integrative GWP prediction model that combines molecular descriptors (MACCS keys and Mordred descriptors) with process information (process title, description, and location) to improve predictive accuracy and interpretability. Using a deep neural network (DNN) model, we achieved an $R^2$ of 86% on test data with Mordred descriptors, process location, and description information, representing a 25% improvement over the previous benchmark of 61%; XAI analysis further highlighted the significant role of process title embeddings in enhancing model predictions.

To enhance interpretability, we employed a Kolmogorov–Arnold Network (KAN) to derive a symbolic formula for GWP prediction, capturing key molecular and process features and providing a transparent, interpretable alternative to black-box models, enabling users to gain insights into the molecular and process factors influencing GWP. Error analysis showed that the model performs reliably in densely populated data ranges, with increased uncertainty for higher GWP values. This analysis allows users to manage prediction uncertainty effectively, supporting data-driven decision-making in chemical and process design. Our results suggest that integrating both molecular and process-level information in GWP prediction models yields substantial gains in accuracy and interpretability, offering a valuable tool for sustainability assessments. Future work may extend this approach to additional environmental impact categories and refine the model to further enhance its predictive reliability.

## Acknowledgement

The authors would like to acknowledge the financial support from the King's College London Net Zero Centre Ph.D. Scholarship scheme.

## 1. Introduction

Life Cycle Assessment (LCA) is an essential tool for quantifying the environmental impacts of products and processes across their full life cycles. Among the various impact categories evaluated in LCA, GWP is especially crucial due to its direct connection to climate change mitigation efforts [1]. However, the rapid emergence of novel chemicals and processes has posed challenges for existing LCA inventory databases, which have become increasingly resource-intensive to develop, frequently delayed in reporting, and often contain data gaps. These limitations obstruct the timely and accurate environmental assessments required for new or rapidly advancing technologies.

To address these challenges, researchers have concentrated on developing predictive models that estimate GWP based on molecular structures and physicochemical properties. Recent studies employing AI models for GWP prediction, summarized in Table 1-1, show that most researchers have utilized the Ecoinvent database to train models using GWP values for organic compounds. Zhu et al. proposed a high-throughput screening framework to identify environmentally favourable chemical substitutes by developing a DNN based model capable of predicting various LCA endpoints [2]. This model exhibited strong predictive performance for the EI99 total and ReCiPe total categories; however, it displayed lower accuracy for the ecosystem endpoint, with $R^2$ values of 65% and 63%. Similarly, Song et al. achieved high predictive accuracy for other categories such as Eco-indicator 99, reaching an $R^2$ of up to 87%, yet their GWP prediction accuracy remained low at 48% [3]. These findings highlight that, despite the high importance of GWP, it remains one of the most challenging LCA categories to predict accurately, underscoring the need for enhanced GWP predictive models. In an attempt to overcome these limitations, Sun et al. applied advanced feature engineering techniques, including a mutual information-permutation importance method and principal component analysis (PCA), which contributed to performance improvements [4]. Nonetheless, most studies have focused solely on Quantitative Structure-Property Relationship (QSPR) modeling, relying exclusively on molecular structural information for GWP predictions. This exclusive focus reveals two critical limitations inherent in existing GWP predictive models.

**Table 1-1 AI-based Global Warming Potential Prediction Studies**

| Authors | Target chemicals | Data size | Features | Model | Performance ($R^2$) |
|---|---|---|---|---|---|
| Zhu et al. [2] | Organic | 224 | Rdkit, AlvaDesc | DNN | 63% |
| Song et al. [3] | Organic | 166 | Dragon 7 | DNN | 48% |
| Sun et al. [4] | Organic | 187 | PaDEL-descriptor | DNN | 81% |
| Kleinekorte et al. [5] | Organic | 500 | Process descriptors, SMILES, graph | GPR, Encoder-decoder | 61% |

The first limitation is the absence of process and locational information in existing models. In LCA, the GWP and similar impact values are fundamentally determined by the cumulative production stages, including the selection of target materials, optimization of reaction pathways, and overall process design and optimization. These stages culminate in measurements of raw material usage, heat, and electricity consumption within the process. Consequently, attempting to predict the final GWP value solely from molecular-level information, which represents the most basic level in LCA, presents inherent limitations. For example, Table 1-2 shows GWP data for ethanol production processes, where the GWP can vary by up to 424% depending on the specific production process. Using only chemical descriptors to predict GWP, however, restricts the model to predicting average GWP values, limiting its practical applicability.

Kleinekorte et al. addressed this issue by proposing a model that embeds not only chemical descriptors but also process-related information, such as the stoichiometric sum of the reactants' impacts [5]. However, as the process information in their study was based on assumptions about the processes, it involved a high degree of uncertainty, resulting in a relatively low predictive performance, with an $R^2$ value of 61%, highlighting the need for further refinement. Additionally, factors such as electricity costs and raw material expenses can vary significantly depending on the location targeted for LCA value estimation, thus greatly impacting the GWP. This emphasizes the lack of locational data integration as another critical shortfall in current models.

**Table 1-2 GWP Variability in Ethanol Production Processes**

| Process Title | Chemical Information | GWP | Average GWP |
|---|---|---|---|
| ethylene hydration \| in 99.7% solution state, from ethylene | CCO | 1.220179 | |
| synthetic fuel production, from coal, high temperature Fisher-Tropsch operations \| in 99.7% solution state, from ethylene | CCO | 6.389138 | |
| dewatering of ethanol from biomass, from 95% to 99.7% solution state \| in 99.7% solution state, from fermentation | CCO | 1.546731 | |
| ethanol production from potatoes \| in 95% solution state, from fermentation | CCO | 2.325355 | |
| ethanol production from rye \| in 95% solution state, from fermentation | CCO | 1.490242 | |
| ethanol production from maize \| in 95% solution state, from fermentation | CCO | 1.261104 | **1.960850** |
| dewatering of ethanol from biomass, from 95% to 99.7% solution state \| in 99.7% solution state, from fermentation | CCO | 1.304751 | |
| market for ethanol, without water, in 99.7% solution state, from ethylene \| in 99.7% solution state, from ethylene | CCO | 1.916034 | |
| ethanol production from rye \| in 95% solution state, from fermentation | CCO | 1.493039 | |
| ethylene hydration \| in 99.7% solution state, from ethylene | CCO | 1.377159 | |
| market for ethanol \| in 99.7% solution state, from ethylene | CCO | 1.245622 | |

The second issue lies in the low interpretability of existing models. While leading studies have used black-box models, such as DNN, they have attempted to maximize interpretability by employing XAI techniques. However, model-agnostic XAI methods, such as SHAP, impose interpretability as a form of post-analysis, which does not address the inherent interpretability limitations of DNNs, which are highly parameter-intensive by nature. This lack of interpretability is a chronic issue not only in the field of chemoinformatics but also across decision-support tasks, prompting active research into enhancing model interpretability directly within the model itself.

In this context, KAN, a recent advancement in computer science, has gained prominence as a powerful alternative to DNNs, combining high interpretability with a compact model architecture [6][7]. KAN is particularly suited to tasks that involve extracting generalized equations from data through pruning, demonstrating significant potential for GWP prediction models, where interpretability is crucial.

In conclusion, GWP prediction models that utilize AI must meet two essential requirements:

1. A predictive model that can integrate both chemical and process information,
2. A model that aligns with domain knowledge, providing explainable predictions.

In this study, we propose a comprehensive GWP prediction model that incorporates chemical structure, physicochemical properties, production process information, and regional context. By addressing the factors previously overlooked, our primary objective is to improve the accuracy and reliability of GWP predictions. Additionally, we develop a white-box GWP prediction model based on the highly interpretable KAN, facilitating more informed decision-making in sustainable chemical process design.

# 2. Methods

## 2.1 Data Collection and Preprocessing for GWP Modelling

Data for this study were collected using the Ecoinvent v3.8 database, which is widely regarded for its extensive data coverage and reliability [8]. The ReCiPe (H) methodology was utilized as the impact assessment method [9]. In total, 2,544 data entries were collected, encompassing 487 different chemicals. Among these entries, there were 1,114 organic flows and 1,430 inorganic flows, with a higher prevalence of inorganic flows. Although focusing solely on organic flows, as in previous studies, might seem reasonable, we included inorganic flows in the training dataset, as their process and regional information could contribute valuable insights to model learning. However, market data within the dataset were excluded, as these entries represent averaged GWP values across various processes, which could inadvertently provide overly deterministic answers and negatively impact fair model evaluation. Consequently, approximately 855 market data entries were removed, leaving a final dataset of 1,689 GWP entries.

While 90% of the dataset comprises GWP values below 27.85 kg $CO_2$-eq, the mean value stands at 687.20 kg $CO_2$-eq, highlighting substantial data skewness. To address this imbalance, we applied a log transformation, which successfully mitigated the effect of excessively high LCA values, as shown in Fig. 2-1(a) [10]. Instead of relying solely on standard scaling or variance-based preprocessing, we evaluated model performance in relation to dataset size and scale to establish a robust preprocessing standard, as outlined in Section 3.1.

Regional information is also essential in determining GWP, though the original data included partial locational data in a country-city format. Since this data format was insufficient for effective training and risked introducing noise, we consolidated city-level data to a country-level label. The regional distribution of refined chemical flows is illustrated in Fig. 2-1(b). This donut chart presents the regional distribution of GWP data within the dataset, with broader regions such as "Rest-of-World" (30.4%), "Global" (23.3%), and "Europe" (21.4%) contributing the largest shares. Among individual countries, China, Switzerland, and Canada provide the most data, with contributions of 3.9%, 2.7%, and 2.5%, respectively. This distribution supports a well-balanced dataset with meaningful representation across global regions and specific countries, facilitating a comprehensive GWP assessment.

The Ecoinvent database provides both a process title, which offers a brief process description, and a more detailed process description for each entry. Table 1-2 lists the process titles, and Fig. 3-3(a) displays examples of process descriptions, which provide a detailed context beyond the process title. Without these distinctions, it would be impossible to differentiate identical substances by molecular characteristics alone. Therefore, we also collected additional data on process titles and descriptions, which capture information on process, reaction, and concentration characteristics.

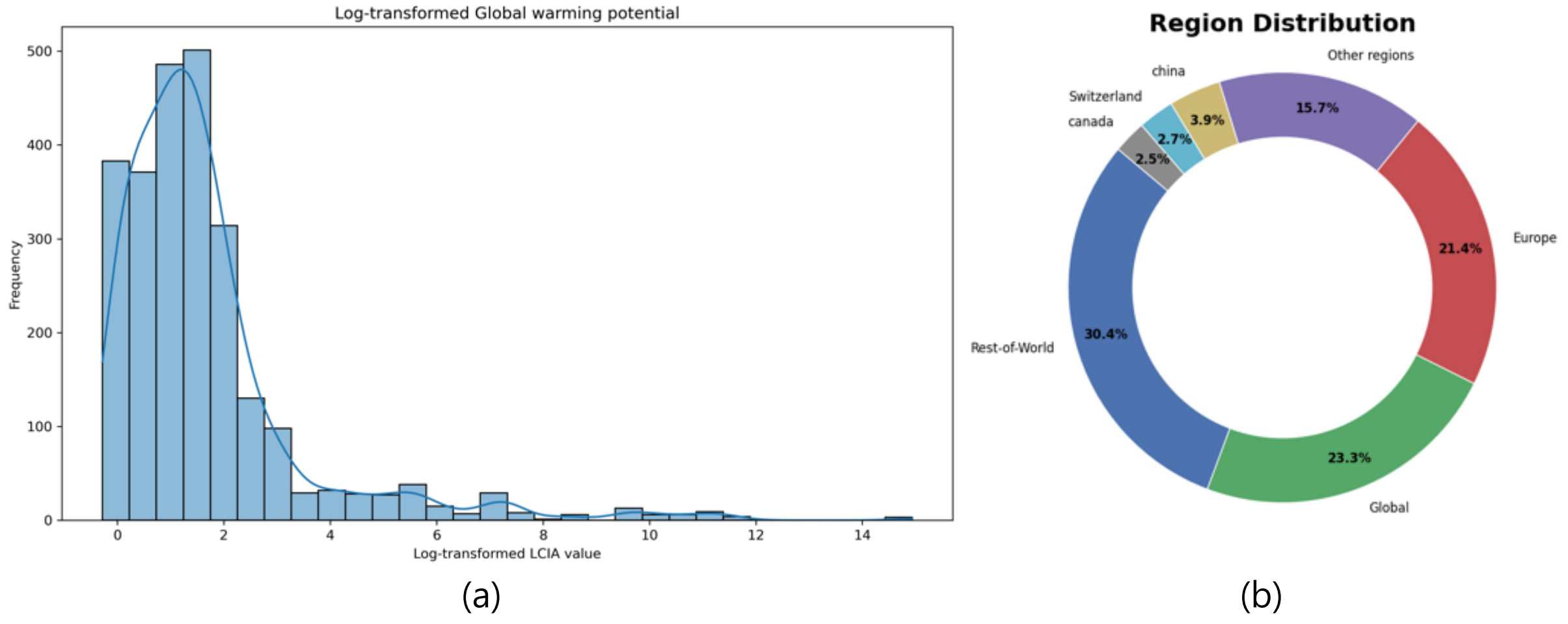

**Fig. 2-1 (a) Distribution of Log-transformed GWP values. (b) Regional Distribution of Chemical Flows.**

## 2.2 Integrative Feature Engineering with Molecular and Process Descriptors

In chemoinformatics, where domain expertise is paramount, the careful selection and engineering of features are essential for model performance [11]. Quantitative Structure-Activity Relationship (QSAR) modelling, a core task within this field, exemplifies the significance of feature selection, with tools such as RDKit and PaDEL providing diverse descriptors that continue to inspire the development of new feature sets [12]. Feature extraction methods like PCA and feature elimination techniques are standard practices, ensuring that only the most relevant features are retained for model optimization [13].

For our study, we incorporated two distinct types of descriptors to capture both structural simplicity and molecular complexity. We selected the MACCS keys, a set of 166 binary features representing fundamental molecular functional groups, and the Mordred descriptors, which provide detailed molecular structural and physicochemical properties [14]. These descriptors allow us to model tasks that range from simple interpretations of molecular structure to more complex combinations of physicochemical features, providing a balanced representation of molecular information for the model.

To enhance model interpretability without compressing critical features, we employed Recursive Feature Elimination with Cross-Validation (RFECV) to identify and retain the most influential features. This approach allowed us to systematically eliminate features with minimal impact on model performance, resulting in an optimized feature set tailored for our analysis.

A unique aspect of this study is the inclusion of process data as input features for GWP prediction. Given the availability of process titles, descriptions, and location information, we performed text embeddings on these features. Text embedding is inherently data-intensive, and producing a reliable embedding model with a limited dataset of around 1,000 entries is challenging. To address this, we utilized OpenAI's pre-trained text-embedding-3-small model, a transformer-based large language model with over a trillion parameters, to embed each of the process title, description, and location into 1,536-dimensional vectors.

However, embedding these three features results in 4,608 dimensions, which is substantially higher than the number of samples (1,689), potentially leading to the "curse of dimensionality." To mitigate this, we applied PCA to reduce the dimensionality and experimented with different dimensional configurations to identify the optimal balance between model performance and feature dimensionality.

Lastly, to rigorously test our hypothesis that incorporating process information could enhance GWP prediction accuracy, we evaluated the predictive power of individual descriptors. We then assessed model performance across various feature combinations, including chemical descriptors, process titles, process descriptions, and process locations. This systematic evaluation allowed us to determine the most effective feature combinations for improving model accuracy, thereby confirming the value of integrating both molecular and process-level data for GWP prediction.

## 2.3 Modelling Framework for GWP Prediction

An overview of our proposed GWP prediction model is shown in Fig. 2-2, with a DNN model used as a representative example for intuitive understanding. First, we converted the chemical names in the collected GWP dataset into SMILES notation, which allowed us to utilize MACCS keys and Mordred descriptors to extract chemical fingerprints and physicochemical properties. Additionally, we embedded process-related information, including process titles, descriptions, and locations, using the text-embedding-small model.

The resulting chemical features and latent vectors from the process data were then used as input features for training the GWP prediction model. For benchmarking, we selected fundamental models frequently applied in LCA prediction and QSAR modelling, including Random Forest (RF) [15], XGBoost [16], and DNN [17]. Finally, we included the KAN model due to its high interpretability, to compare its performance with that of other models in GWP prediction.

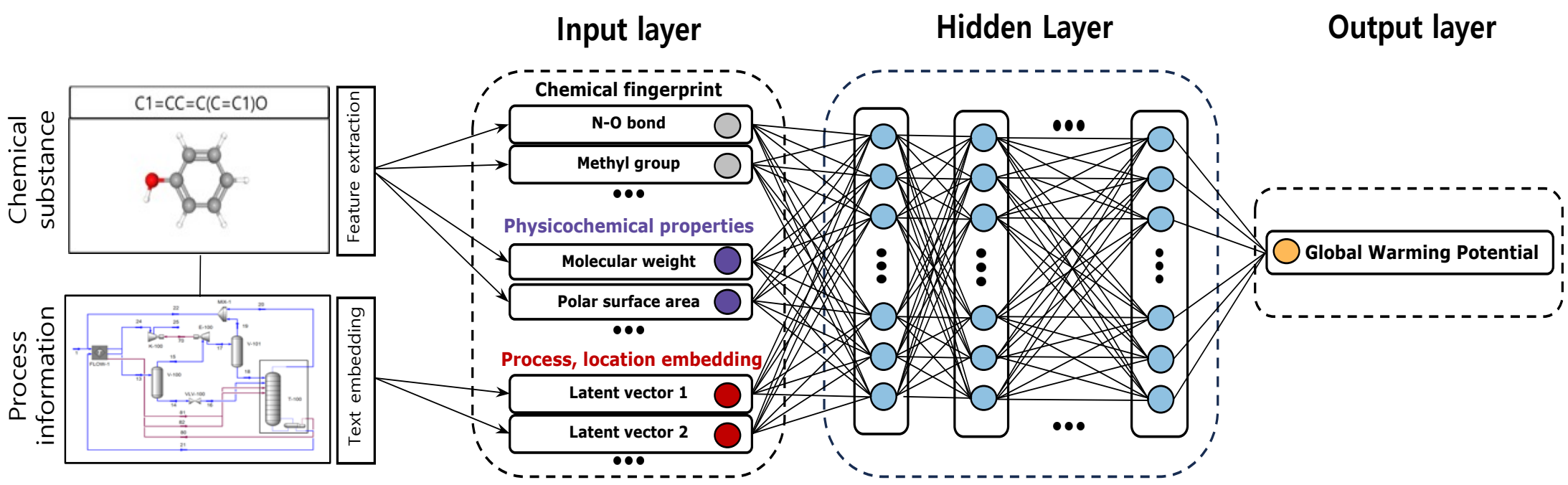

**Fig. 2-2 Overview of the proposed GWP prediction model.**

# 3. Results and Discussion

## 3.1 Learning Curve Analysis

To determine the optimal data volume for effective training, we conducted a learning curve analysis, considering the high variance and scale differences within the dataset. We partitioned the entire dataset into 10 segments based on a logarithmic scale. By incrementally adding one partition at a time, we progressively increased the data scale and evaluated the performance of the XGBoost model at each stage.

As shown in Fig. 3-1, the highest $R^2$ score was achieved with Fold 2, which utilized 87.4% of the available data and yielded an $R^2$ of 82.7%. Notably, a sharp performance decline was observed after Fold 8, where GWP values exceeded 147,266, resulting in a substantial drop in the $R^2$ score and indicating a decrease in model effectiveness.

To balance model performance, scalability, and robustness, we limited the training data to Fold 8, capturing the largest feasible dataset for effective training without compromising model stability.

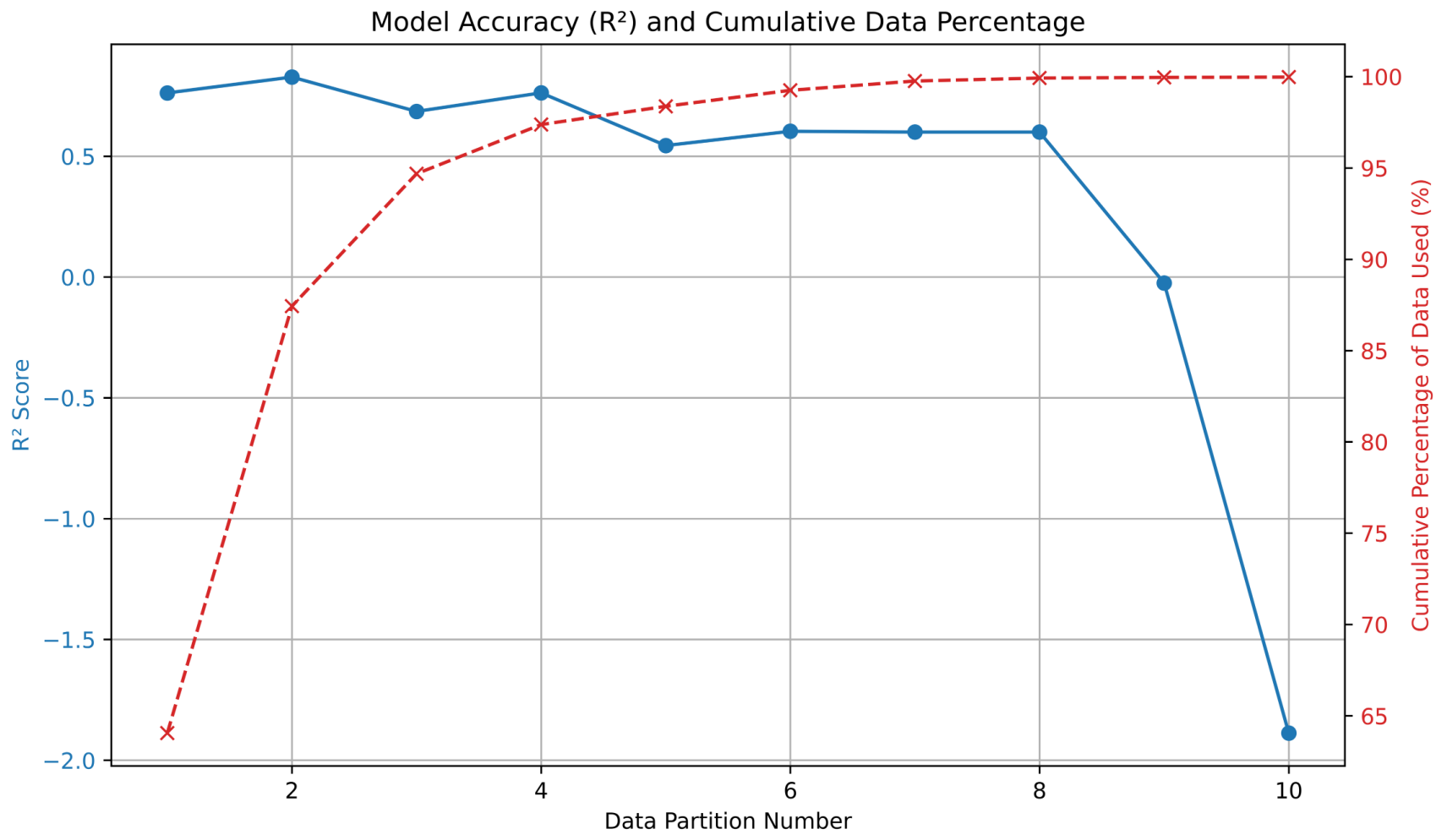

**Fig. 3-1 Learning curve analysis depicting the $R^2$ score and cumulative data usage percentage across data partitions.**

## 3.2 Dimensionality Reduction in Chemical Feature Engineering

Before incorporating process information into the model, we first reduced the dimensionality of the chemical features to ensure efficient model development. We employed RFECV to iteratively remove low-importance features from the combined set of MACCS keys and Mordred descriptors. This process involved training the model after each feature removal and assessing its performance. For this comparison, we used the XGBoost model due to its high training efficiency, performing 10-fold cross-validation and evaluating performance based on negative Mean Squared Error (MSE).

As shown in Fig. 3-2, model performance improved sharply as the feature count increased from 1 to 20. However, this improvement plateaued beyond 20 features, with minimal changes observed after 55 features. Based on these results, we selected a reduced feature set of 55 features, balancing interpretability and performance without significant loss in accuracy. This reduction allowed us to condense the total feature count from 1,991 to 55, a substantial simplification that preserves model effectiveness.

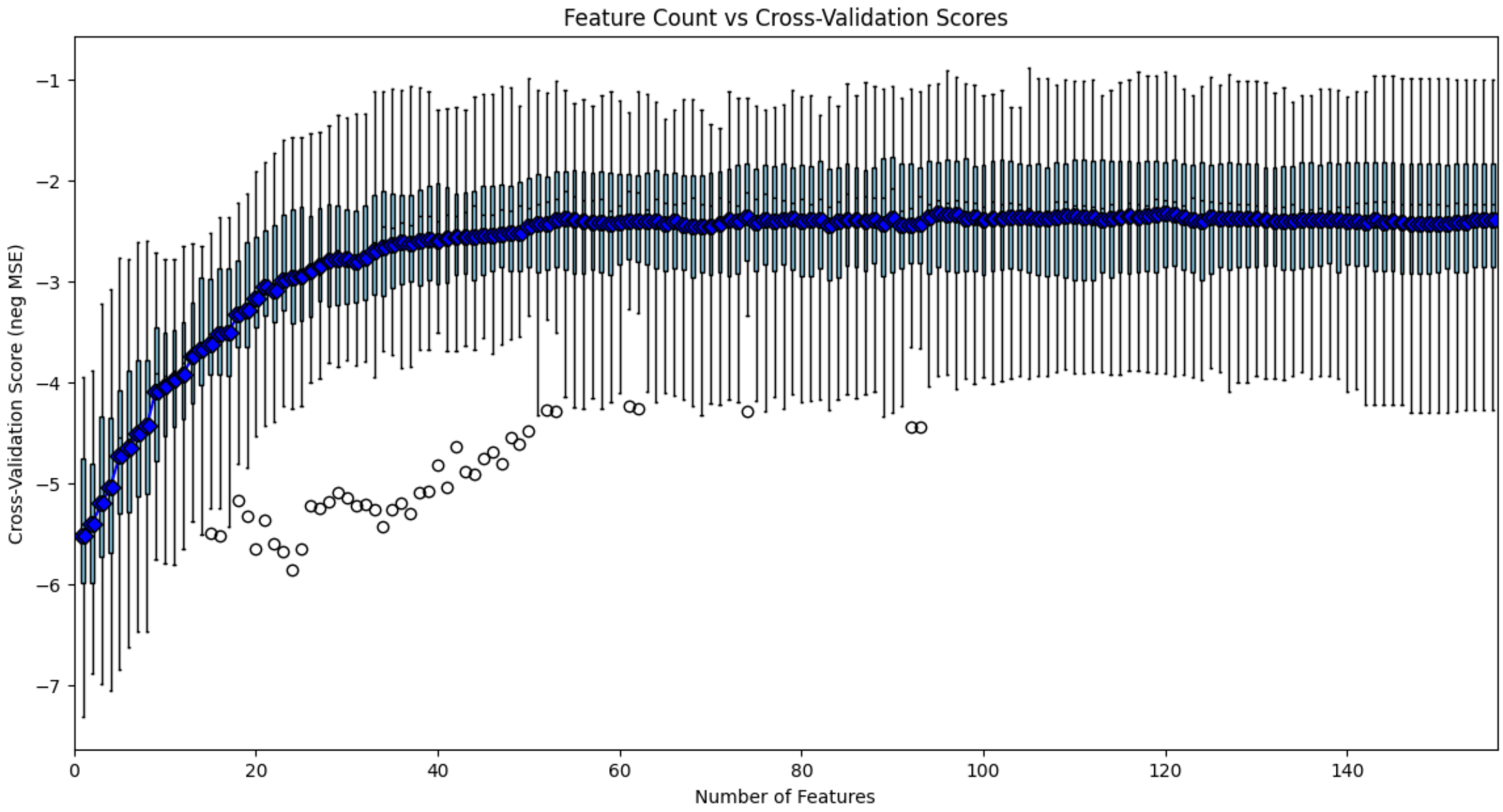

**Fig. 3-2 Impact of feature count on cross-validation performance.**

## 3.3 Embedding Process and Regional Information Using a Pre-trained Large Language Model

We utilized OpenAI's text-embedding-3-small model to embed process titles, descriptions, and locations (as illustrated in Fig. 3-3(a)) into a 1,536-dimensional vector. To reduce dimensionality, we applied PCA. However, solely relying on eigenvalues or cumulative variance to select dimensions carries a degree of uncertainty. Thus, although time-intensive, we incrementally increased the dimensionality to identify the optimal embedding dimensions based on model performance.

To evaluate performance, we conducted 10-fold cross-validation using the XGBoost model, with $R^2$ as the performance metric. As shown in Fig. 3-3(b), we identified 40 dimensions as optimal for the process title, which contained relatively less information, and 60 dimensions for the process description, which held more detailed data. Beyond these dimensions, further increases in dimensionality led to a decline in model performance, likely due to additional noise from unnecessary dimensions interfering with model learning.

| Process Description |
|---|
| Production of (sulfonyl)urea-compounds including materials, energy uses, infrastructure and emissions. It includes all upstream activities ("from cradle") related to the productio n process. Production of (sulfonyl)urea-compounds including materials, energy uses, inf rastructure and emissions. |

| Process Title |
|---|
| ethylene hydration \| in 99.7% solution state, from ethylene |
| synthetic fuel production, from coal, high temperature Fisher-Tropsch operations \| in 99.7% solution state, from ethylene |
| dewatering of ethanol from biomass, from 95% to 99.7% solution state \| in 99.7% solution state, from fermentation |
| ethanol production from potatoes \| in 95% solution state, from fermentation |
| ethanol production from rye \| in 95% solution state, from fermentation |
| ethanol production from maize \| in 95% solution state, from fermentation |
| dewatering of ethanol from biomass, from 95% to 99.7% solution state \| in 99.7% solution state, from fermentation |
| ethanol production from rye \| in 95% solution state, from fermentation |
| ethylene hydration \| in 99.7% solution state, from ethylene |

(a)

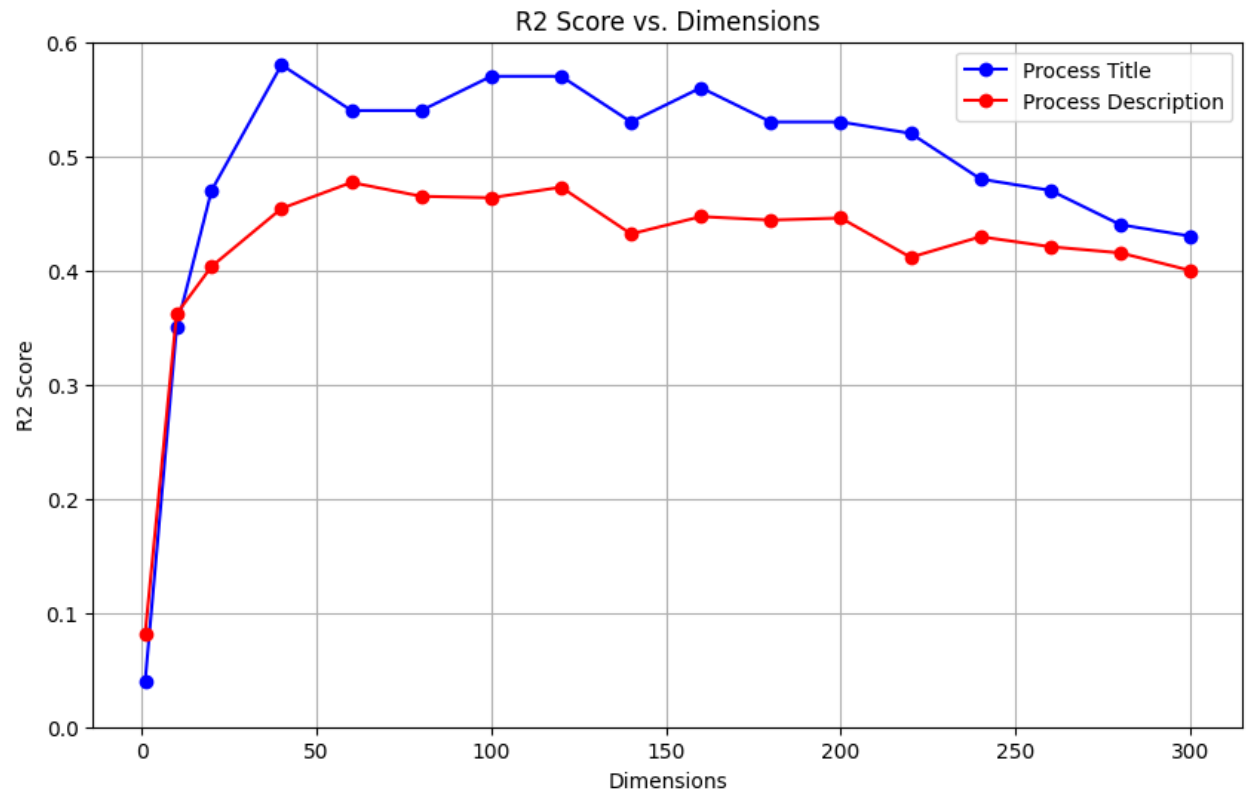

(b)

**Fig. 3-3 Evaluation of $R^2$ score versus dimensionality for process titles and descriptions.**

To verify the distribution of embedded data in the latent space, we used Uniform Manifold Approximation and Projection (UMAP) to visualize the latent vectors in two dimensions [18]. Fig. 3-4 shows the embedding visualization for ethanol production processes only. As depicted, fermentation-based ethanol production processes (blue points) and sugarcane-based ethanol processes (green points) are effectively separated, indicating that the pre-trained embedding model successfully distinguishes between different process types.

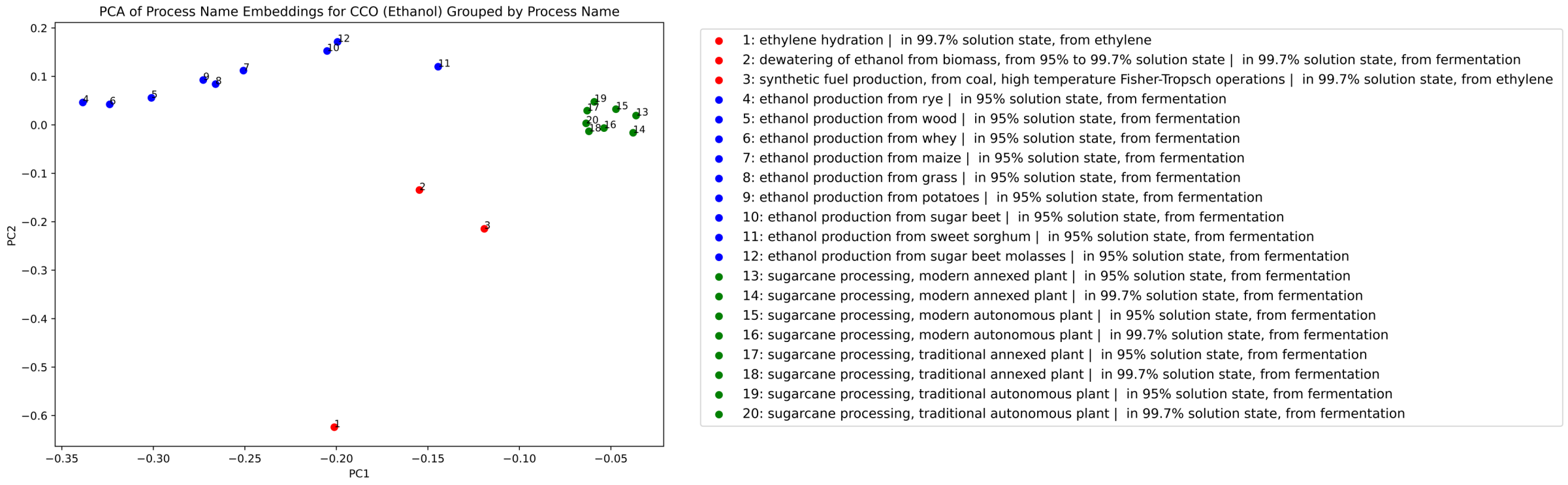

**Fig. 3-4 UMAP visualization of ethanol production process embeddings.**

For process location data, which does not exhibit large variance, we determined that high dimensionality was unnecessary and reduced it to 10 dimensions. The two-dimensional UMAP visualization in Fig. 3-5 shows that broader categories such as "Rest-of-World," "Global," and "Europe" are distributed further from country-level names, facilitating more efficient model learning.

| Process location |
|---|
| Brazil |
| Europe |
| Europe without Switzerland |
| China |
| Rest-of-World |
| Global |

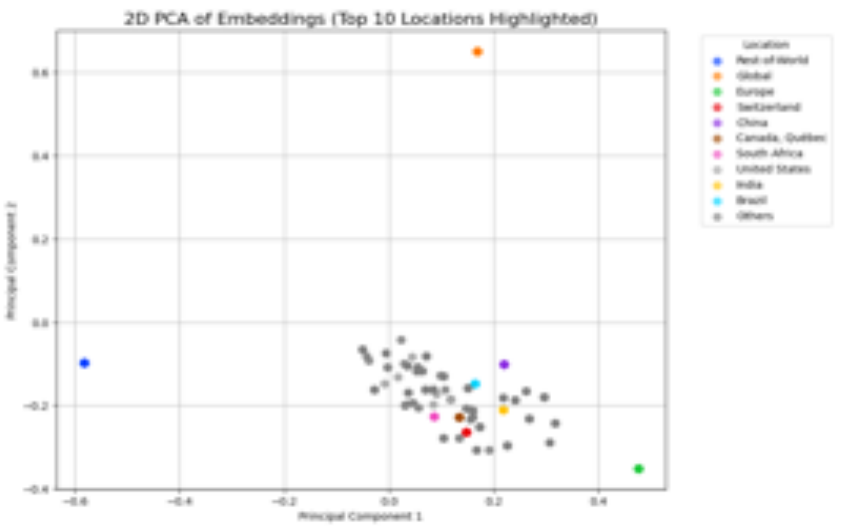

**Fig. 3-5 UMAP visualization of process location embeddings.**

## 3.4 Model Benchmarking and Feature Impact on GWP Prediction

We compared the performance of four models—RF, XGBoost, DNN and KAN—using both MACCS keys and Mordred descriptors as chemical features. Table 3-1 shows the $R^2$ prediction results for each model with different feature combinations. The Mordred descriptors outperformed the MACCS keys across all models, with the DNN model demonstrating an 11% performance improvement when using Mordred descriptors. This suggests that the physicochemical properties captured by the Mordred descriptors have a greater impact on GWP prediction compared to the structural information provided by MACCS keys. This finding underscores the relevance of physicochemical properties in GWP modelling.

Given the superior performance of the Mordred descriptors, we selected them as the primary chemical feature and examined their combinations with process-related features, including process location, process title, and process description. The results indicate that the combination of Mordred descriptors with process location and process description yielded the highest performance across all models. Notably, using only the process description alongside the chemical features improved prediction accuracy by up to 9% compared to using chemical features alone.

In terms of model ranking based on performance, the DNN model was the best, followed by XGBoost, KAN, and RF. Despite dimensionality reduction, the feature set remained high-dimensional (165 features), favouring the learning capacity of the DNN model. Interestingly, the tasks where KAN outperformed the DNN often involved fewer than 30 features, suggesting that KAN struggle to handle high-dimensional tasks and the redundancy in chemical features [19][20][21][22]. However, the text embedding feature set, with lower redundancy, resulted in comparable performance, with the DNN outperforming by only 3%.

The highest-performing model used the DNN trained on a combination of Mordred descriptors, process location, and process description, achieving an $R^2$ of 86%. This is approximately 25% higher than the current benchmark of 61% achieved by Kleinekorte et al. on a similar dataset and task. Furthermore, our model outperformed traditional GWP prediction models for production materials by a margin of 5% to 26%, even for the more challenging task of predicting GWP based on both production material and process information.

**Table 3-1 Comparison of model prediction performance ($R^2$) across different feature combinations.**

| | Ma | Mo | L | T | D | Mo+L | Mo+T | Mo+D | L+T | L+D | T+D | L+T+D | Mo+L+T | Mo+L+D | Mo+T+D | Mo+L+T+D |
|---|---|---|---|---|---|---|---|---|---|---|---|---|---|---|---|---|
| RF | 54% | 55% | 9% | 42% | 58% | 53% | 46% | 61% | 42% | 57% | 58% | 58% | 46% | 61% | 61% | 60% |
| XGB | 64% | 68% | 8% | 50% | 77% | 63% | 68% | 80% | 51% | 81% | 77% | 79% | 68% | 83% | 78% | 78% |
| DNN | 60% | 71% | 8% | 60% | 76% | 62% | 78% | 80% | 56% | 78% | 79% | 78% | 78% | 86% | 79% | 81% |
| KAN | 57% | 66% | 7% | 51% | 75% | 53% | 76% | 61% | 53% | 74% | 78% | 74% | 75% | 78% | 56% | 76% |

[1]

## 3.5 XAI and Symbolic Model Analysis

To identify the primary factors contributing to GWP prediction, we conducted a post-hoc analysis using XAI techniques on the XGBoost model, which achieved the highest predictive performance with MACCS keys. The results, illustrated in Fig. 3-6, reveal that the presence of halogen atoms, oxygen, and unusual elements—excluding basic elements such as H, C, N, O, Si, P, S, F, Cl, Br, and I—as well as chlorine and methyl groups, contributed most significantly to GWP predictions.

Halogens like chlorine tend to persist in the atmosphere, enhancing greenhouse effects. Functional groups containing oxygen exhibit polarity, which affects the physicochemical properties of molecules. These characteristics, in turn, influence molecular absorption, emission properties, and reactivity, thereby indirectly impacting GWP. Structures based on nitrogen, such as amines or other nitrogen compounds, can also affect molecular stability and degradation rates. Nitrogen compounds that are resistant to atmospheric breakdown may persist, potentially contributing to greenhouse effects over prolonged periods. These findings align with established domain knowledge regarding GWP, validating the model's predictive insights.

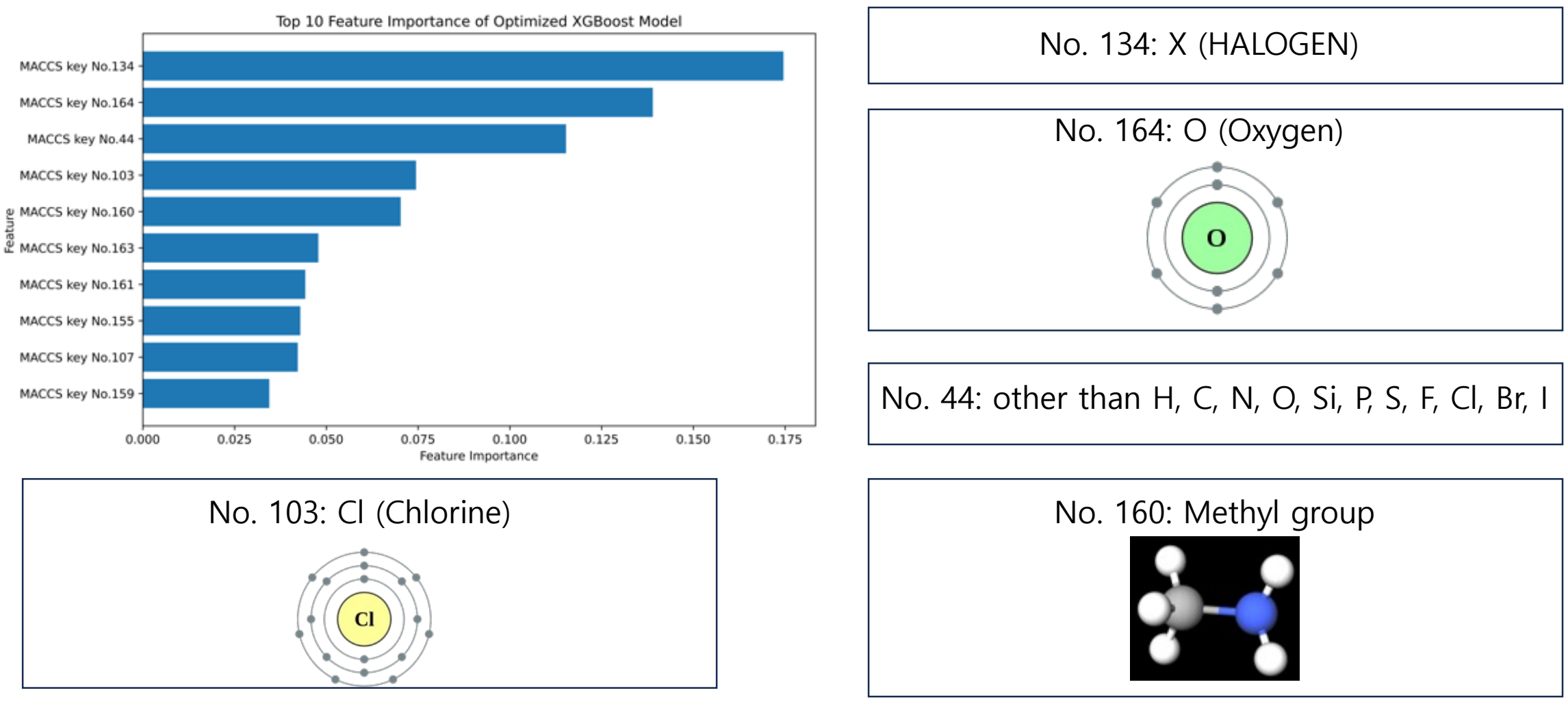

**Fig. 3-6 Feature importance analysis of XGBoost model for GWP prediction.**

[1] Ma: MACCS keys Mo: Mordred descriptors L: Process Location T: Process Title D: Process Description

Next, we examined the overall contribution of each feature. As shown in Fig. 3-7, the feature importance analysis reveals that the vector embedding of the process title plays a significant role in the model, contributing the most among all features. This suggests that process title information provides valuable context for GWP prediction, likely due to its ability to capture specific aspects of the process that are directly related to environmental impact. Other features, including structural and descriptive information, also contribute meaningfully, indicating a balanced input from both chemical and process-related data.

The pie chart on the right summarizes the overall feature contributions by category: structural information (50.4%), title information (22.7%), description information (22.6%), and location information (4.3%). This distribution suggests that chemical and process information contribute approximately equally to model training, reaffirming that process information is as crucial as chemical structure information in GWP prediction. This balanced contribution likely underpins the substantial improvement in model performance.

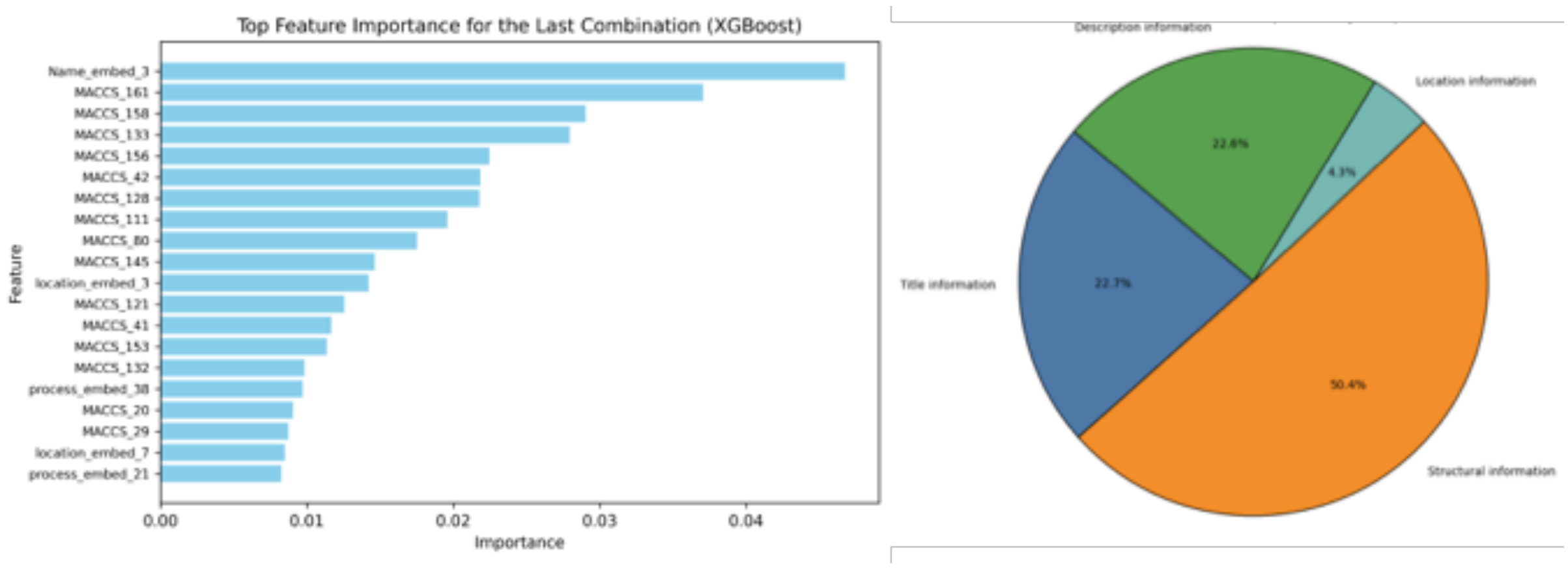

**Fig. 3-7 Feature importance analysis of GWP prediction model.**

Finally, we leveraged the symbolic distillation capability of the KAN, one of its most powerful features, to derive a highly generalized and interpretable formula from the model trained on Mordred descriptors. Each feature in the resulting formula (1) contributes to prediction accuracy by encapsulating essential aspects of molecular structure and chemical properties that impact GWP.

$$(1)\ GWP = \frac{4919 \cdot \mathrm{SpMax_A} + 105597 \cdot \mathrm{nAromAtom} - 19339 \cdot \mathrm{nAromBond} + 112143 \cdot \mathrm{nC}}{1000000} + \frac{4713 \cdot \mathrm{ATS5dv} + 12171 \cdot \mathrm{ATS6dv} + 24773 \cdot \mathrm{SpAD_A} - 39567 \cdot \mathrm{SpAbs_A}}{1000000} - \frac{94409 \cdot \exp\left(-\frac{1141 \cdot \mathrm{VE3_A}}{1000000} + \frac{\mathrm{nAtom}}{240} - \frac{1777 \cdot \mathrm{nBase}}{500000}\right)}{1000000}$$

The symbolic formula extracted from the KAN highlights several key features that contribute to GWP prediction by encapsulating essential aspects of molecular structure and chemical properties. Spatial Maximum Autocorrelation for atoms (SpMax_A) captures the spatial autocorrelation between atoms within a molecule, reflecting structural complexity and interatomic interactions; higher atomic arrangement complexity correlates with greater thermal stability, making the molecule more resistant to degradation and thus likely to contribute to GWP. Aromatic features, such as the Number of Aromatic Atoms (nAromAtom) and Number of Aromatic Bonds (nAromBond), indicate aromaticity, a characteristic associated with stability. For example, benzene rings, due to their resistance to decomposition, enhance atmospheric persistence and subsequently impact GWP. Carbon atom count (nC) is also critical, as it directly correlates with molecular size and weight, attributes that generally increase a molecule's resistance to degradation. Features such as Autocorrelation of Distance Vectors (ATS5dv and ATS6dv) measure interactions at specific distance intervals, representing three-dimensional structural properties that suggest stability and extended atmospheric persistence, both contributing to higher GWP. The Spatial Average and Absolute Distance Autocorrelation for atoms (SpAD_A and SpAbs_A) features capture the overall structural complexity by reflecting average and absolute interatomic autocorrelation across the molecule. Generally, complex and stable structures are more resistant to degradation, further impacting GWP. The 3D Bertz Complexity Index (VE3_A) provides a quantitative measure of the three-dimensional structural complexity, with higher values indicating more intricate molecular arrangements that are typically associated with increased GWP due to resilience against degradation. Additional features like total atom count (nAtom) and number of basic atoms (nBase) highlight the molecule's scale and potential reactivity, respectively. Larger molecules, often more stable, contribute to prolonged atmospheric presence, while basic atoms can affect reactivity patterns, enhancing atmospheric persistence and GWP impact. By encompassing these features, the formula systematically generalizes domain knowledge related to chemical and structural factors influencing GWP. Each feature effectively captures critical molecular properties, such as structural complexity, aromaticity, stability, and reactivity, aligning with established chemical principles and presenting a highly interpretable, data-driven approach to assessing environmental impact.

Using this symbolic model, we obtained the prediction results shown in Fig. 3-8. The model achieved an $R^2$ accuracy of 59% on the training data and 42% on the test data. Although this accuracy is lower than that of other models trained with more features, it remains acceptable, given that the model is a white-box and interpretable mathematical model. Additionally, since the GWP dataset has considerable variance, we analysed the error across different scales, as shown in Fig. 3-9. In the 0–2 range, where data density is highest, the model exhibited very low error. However, beyond a log-transformed scale of 5, both the error and its variance increased sharply. Based on this reliability assessment, we believe that actual users can predict and manage uncertainty effectively, supporting informed decision-making.

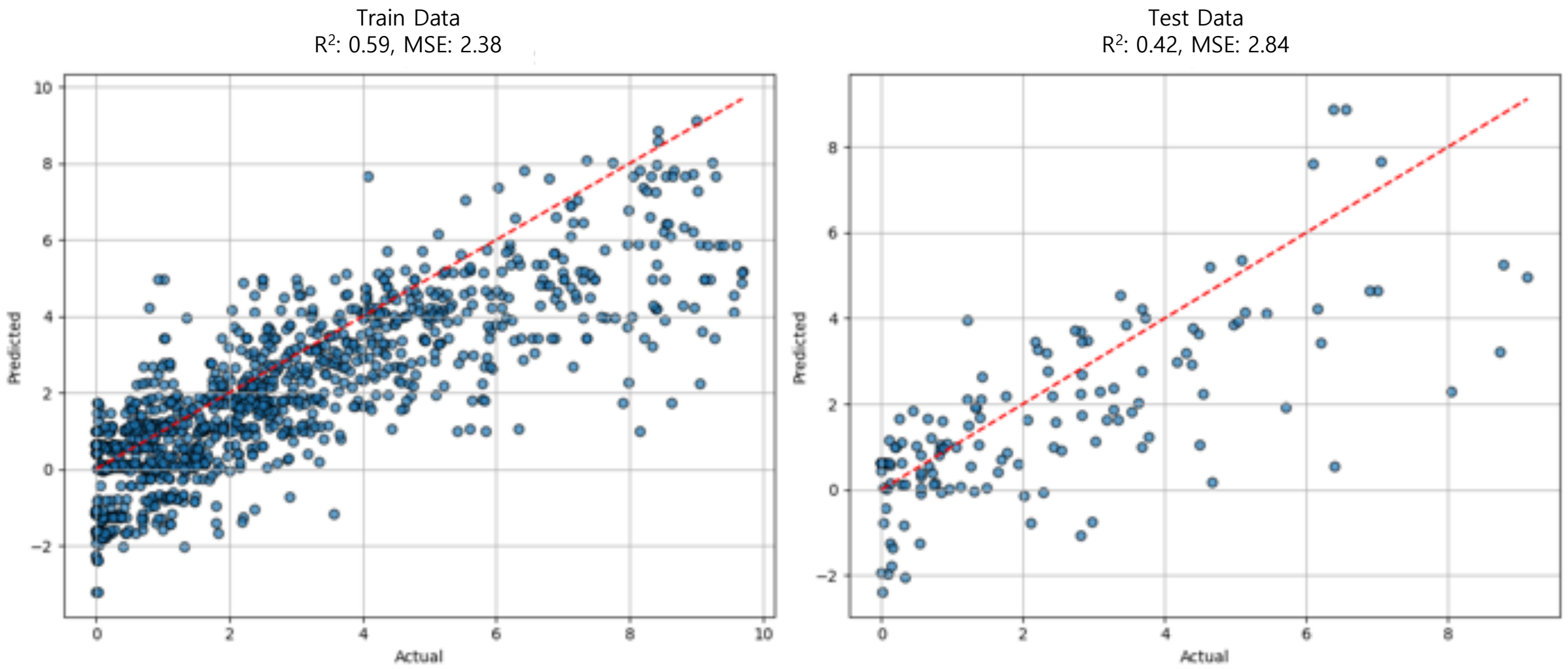

**Fig. 3-8 Prediction Results of the Symbolic Model for GWP Prediction.**

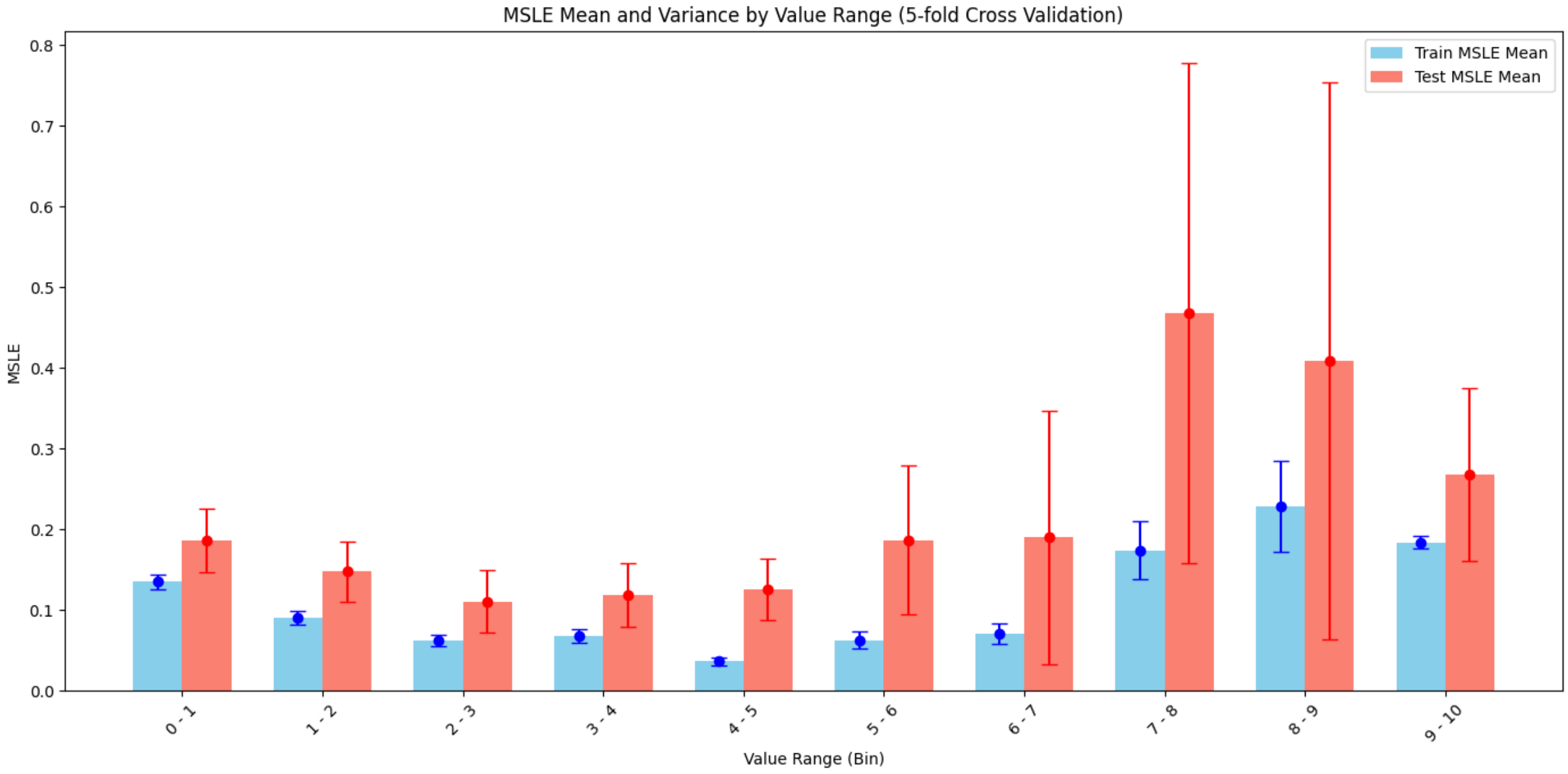

**Fig. 3-9 Error Analysis Across Log-Transformed GWP Scales.**

# 4. Conclusion

In this study, we developed and evaluated a GWP prediction model that integrates both chemical and process-related information, leveraging feature engineering methods, interpretable modelling techniques, and XAI analysis. By combining MACCS keys and Mordred descriptors as molecular features and embedding process information—including title, description, and location—into high-dimensional latent space, we created a comprehensive feature set that significantly improved model performance compared to traditional GWP prediction models relying solely on chemical structure.

Our analysis confirmed the critical role of physicochemical properties captured by Mordred descriptors in accurately predicting GWP, with the DNN model achieving an $R^2$ of 86% on the test set when combining Mordred descriptors with process location and description information. This performance represents an approximate 25% improvement over the existing benchmark model. Through XAI and feature importance analysis, we demonstrated that process title embeddings, which encapsulate contextual information about production processes, contributed substantially to the model's predictions. This finding underscores the importance of process-related data in environmental impact modelling, revealing that process information can be as influential as molecular structure for GWP prediction.

While the DNN model achieved the highest overall accuracy, we also explored the KAN to develop a white-box symbolic model that provides interpretable predictions. Despite achieving a lower $R^2$ accuracy of 59% on the training data and 42% on the test data, the symbolic model's ability to express GWP prediction as a mathematical formula makes it a valuable tool for decision-making, offering transparency and insights into the molecular and process characteristics driving GWP.

Our error analysis revealed that the model performs reliably within the data's densest ranges (0–2 log-transformed scale), with low error rates. However, prediction uncertainty increases sharply for extreme values beyond a log-transformed scale of 5, indicating areas where the model's reliability diminishes. This insight allows potential users to better manage uncertainty when applying the model to decision-making processes, especially in cases involving novel or untested chemicals and processes.

In conclusion, this research demonstrates that integrating molecular and process-level features in GWP prediction models not only improves predictive accuracy but also provides interpretable insights into the factors affecting environmental impact. Our work suggests a pathway toward more reliable, interpretable, and data-driven tools for sustainability assessments, aiding researchers and industry professionals in making informed, environmentally conscious choices in chemical and process design. Future studies may expand upon this approach by incorporating additional environmental impact categories or refining the model to further reduce uncertainty in GWP predictions.